\documentclass[10pt,twocolumn,letterpaper]{article}

\usepackage{cvpr}
\usepackage{times}
\usepackage{epsfig}
\usepackage{graphicx}
\usepackage{amsmath}
\usepackage{amssymb}
\usepackage{multirow}
\usepackage{epsfig}

\usepackage [english]{babel}
\usepackage [autostyle, english = american]{csquotes}
\MakeOuterQuote{"}

\graphicspath{{./images/}}

\usepackage[pagebackref=true,breaklinks=true,letterpaper=true,colorlinks,bookmarks=false]{hyperref}
\hypersetup{citecolor=red} %Cites red (evil)
\hypersetup{linkcolor=dgreen} %Local refs green (good)
\hypersetup{pdftitle={Learning by Association - A versatile semi-supervised training method for neural networks},
           pdfauthor={Philip Haeusser, Alexander Mordvintsev, Daniel Cremers},
            pdfsubject={CVPR 2017 Submission (arXiv.org version)},
            pdfkeywords={deep learning, semi-supervised, training, classification, domain              adaptation}}

\usepackage[usenames,dvipsnames]{xcolor}                                   
\definecolor{dgreen}{rgb}{0.0,0.6,0.0}
\definecolor{dred}{rgb}{0.6,0.0,0.0}
\definecolor{philipcolor}{rgb}{0,0.5,0}                                    
\definecolor{grey}{rgb}{0.6,0.6,0.6}                                       
\newcommand*{\defeq}{\mathrel{\vcenter{\baselineskip0.5ex \lineskiplimit0pt
                     \hbox{\scriptsize.}\hbox{\scriptsize.}}}%
                     =}
\newcommand{\specialcell}[2][c]{%
  \begin{tabular}[#1]{@{}c@{}}#2\end{tabular}}
\usepackage[pagebackref=true,breaklinks=true,letterpaper=true,colorlinks,bookmarks=false]{hyperref}
\addto\extrasenglish{%
}

\cvprfinalcopy % *** Uncomment this line for the final submission

\def\cvprPaperID{38}

\begin{document}

%%%%%%%%% TITLE
\title{Learning by Association\\[1mm]
A versatile semi-supervised training method for neural networks}

\author{Philip Haeusser$^{1,2}$\\
% For a paper whose authors are all at the same institution,
% omit the following lines up until the closing ``}''.
% Additional authors and addresses can be added with ``\and'',
% just like the second author.
% To save space, use either the email address or home page, not both
\and
Alexander Mordvintsev$^2$\\
\and
Daniel Cremers$^1$
\and
$^1$Dept. of Informatics, TU Munich\\
{\tt\small \{haeusser, cremers\}@in.tum.de}
\and
$^2$Google, Inc.\\
{\tt\small moralex@google.com}
}
\maketitle
\thispagestyle{empty}

%%%%%%%%% ABSTRACT
\begin{abstract}
In many real-world scenarios, labeled data for a specific machine learning task is costly to obtain. Semi-supervised training methods make use of abundantly available unlabeled data and a smaller number of labeled examples.
We propose a new framework for semi-supervised training of deep neural networks inspired by learning in humans. "Associations" are made from embeddings of labeled samples to those of unlabeled ones and back. The optimization schedule encourages correct association cycles that end up at the same class from which the association was started and penalizes wrong associations ending at a different class.
The implementation is easy to use and can be added to any existing end-to-end training setup.
We demonstrate the capabilities of learning by association on several data sets and show that it can improve performance on classification tasks tremendously by making use of additionally available unlabeled data. In particular, for cases with few labeled data, our training scheme outperforms the current state of the art on SVHN.
%We also show how to apply this training scheme to the task of domain adaptation, surpassing current state-of-the-art results.
\end{abstract}

%%%%%%%%% BODY TEXT
%%%%%%%%% INTRO
\section{Introduction}
A child is able to learn new concepts quickly and without the need for millions examples that are pointed out individually. Once a child has seen one dog, she or he will be able to recognize other dogs and becomes better at recognition with subsequent exposure to more variety.

In terms of training computers to perform similar tasks, deep neural networks have demonstrated superior performance among machine learning models (\cite{lecun98, alexnet, resnet}).
However, these networks have been trained dramatically differently from a learning child, requiring labels for every training example, following a purely supervised training scheme. Neural networks are defined by huge amounts of parameters to be optimized. Therefore, a plethora of labeled training data is required, which might be costly and time consuming to obtain. 
It is desirable to train machine learning models without labels (unsupervisedly) or with only some fraction of the data labeled (semi-supervisedly).

Recently, efforts have been made to train neural networks in an unsupervised or semi-supervised manner yielding promising results. However, most of these methods require a trick to generate training data, such as sampling patches from an image for context prediction \cite{doersch} or generating surrogate classes \cite{doso14discriminative, pseudo, huang16}. In other cases, semi-supervised training schemes require non trivial additional architectures such as generative adversarial networks \cite{gan} or a decoder part \cite{swwae}.

\begin{figure}[t]
\begin{center}
   \includegraphics[width=1.\linewidth]{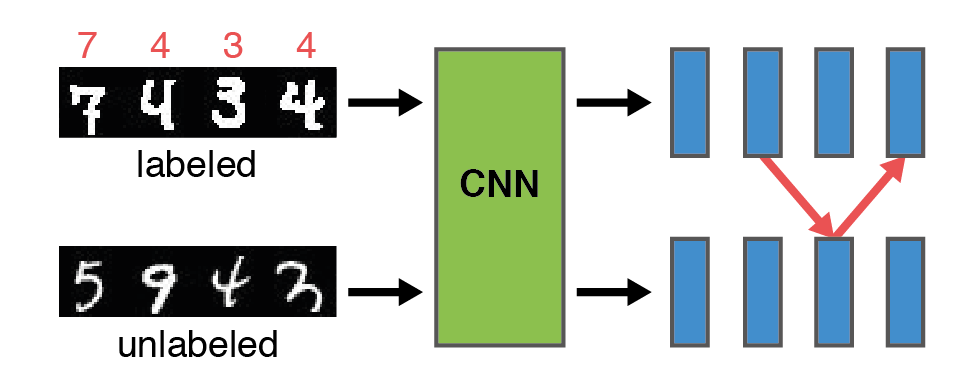}
\end{center}
   \caption{Learning by association. A network (green) is trained to produce embeddings (blue) that have high similarities if belonging to the same class. A differentiable association cycle (red) from embeddings of labeled ($A$) to unlabeled ($B$) data and back is used to evaluate the association.}
\label{fig:teaser}
\end{figure}

We propose a novel training method that follows an intuitive approach: learning by association (\autoref{fig:teaser}).
We feed a batch of labeled and a batch of unlabeled data through a network, producing embeddings for both batches.
Then, an imaginary walker is sent from samples in the labeled batch to samples in the unlabeled batch. The transition follows a probability distribution obtained from the similarity of the respective embeddings which we refer to as an \emph{association}.
In order to evaluate whether the association makes sense, a second step is taken back to the labeled batch - again guided by the similarity between the embeddings. It is now easy to check if the cycle ended at the same class from which it was started. We want to maximize the probability of consistent cycles, i.e., walks that return to the same class. Hence, the network is trained to produce embeddings that capture the essence of the different classes, leveraging unlabeled data. In addition, a classification loss can be specified, encouraging embeddings to generalize to the actual target task.

The association operations are fully differentiable, facilitating end-to-end training of arbitrary network architectures. Any existing classification network can be extended by our customized loss function.

In summary, our key contributions are:
\begin{itemize}
\setlength\itemsep{0em}
\item A novel yet simple training method that allows for semi-supervised end-to-end training of arbitrary network architectures. We name the method "associative learning".
\item An open-source TensorFlow implementation\footnote{\url{https://git.io/vyzrl}} of our method that can be used to train arbitrary network architectures.
\item Extensive experiments demonstrating that the proposed method improves performance by up to 64\% compared to the purely supervised case.
\item Competitive results on MNIST and SVHN, surpassing state of the art for the latter when only a few labeled samples are available.
%\item State-of-the-art results for domain adaptation \newline (SVHN $\rightarrow$ MNIST and Synthetic Digits $\rightarrow$ SVHN).
\end{itemize}

%-------------------------------------------------------------------------
%%%%%%%%% RW
\section{Related Work}
The challenge of harnessing unlabeled data for training of neural networks has been tackled using a variety of different methods. Although this work follows a semi-supervised approach, it is in its motivation also related to purely unsupervised methods.
A third category of related work is constituted by generative approaches.

\subsection{Semi-supervised training}%%%%%%%%%
The semi-supervised training paradigm has not been among the most popular methods for neural networks in the past.
It has been successfully applied to SVMs \cite{tsvm} where unlabeled samples serve as additional regularizers in that decision boundaries are required to have a broad margin also to unlabeled samples.

%In the field of speech recognition, semi-supervised approaches have been adopted for training neural networks \cite{vesely}, \cite{liao}. Most approaches are different from ours in that they don't process both labeled and unlabeled data simultaneously %in an end-to-end fashion, but are rather used separately at different stages of training.

One training scheme applicable to neural nets is to bootstrap the model with additional labeled data obtained from the model's own predictions.
\cite{pseudo} introduce pseudo-labels for unlabeled samples which are simply the class with the maximum predicted probability. Labeled and unlabeled samples are then trained on simultaneously. In combination with a denoising auto-encoder and dropout, this approach yields competitive results on MNIST.

Other methods add an auto-encoder part to an existing network with the goal of enforcing efficient representations (\cite{ranzato2008semi} \cite{weston2012deep} \cite{swwae}).

Recently, \cite{sajjadi2016mutual} introduced a regularization term that uses unlabeled data to push decision boundaries of neural networks to less dense areas of decision space and enforces mutual exclusivity of classes in a classification task. When combined with a cost function that enforces invariance to random transformations as in \cite{sajjadi2016regularization}, state-of-the-art results on various classification tasks can be obtained.

%To the best of our knowledge, there are no end-to-end semi-supervised methods for neural networks that treat unlabeled data equally (i.e., not in a separate auto-encoder branch or exclusively in a pre-training stage, for example).

\subsection{Purely unsupervised training}%%%%%%%%%
Unsupervised training is obviously more general than semi-supervised approaches. It is, however, important to differentiate the exact purpose. While semi-supervised training allows for a certain degree of guidance as to what the network learns, the usefulness of unsupervised methods highly depends on the design of an appropriate cost function and balanced data sets. For exploratory purposes, it might be desirable that representations become more fine grained for different suptypes of one class in the data set.
Conversely, if the ultimate goal is classification, invariance to this very phenomenon might be more preferable.

\cite{rbmhinton} propose to use Restricted Boltzmann Machines (\cite{rbm}) to pre-train a network layer-wise with unlabeled data in an auto-encoder fashion.

\cite{higgins}\cite{quoc}\cite{swwae} build a neural network upon an auto-encoder that acts as a regularizer and encourages representations that capture the essence of the input.

A whole new category of unsupervised training is to generate surrogate labels from data. \cite{huang16} employ clustering methods that produce weak labels.

\cite{doso14discriminative} generate surrogate classes from transformed samples from the data set. These transformations have hand-tuned parameters making it non-trivial to ensure they are capable of representing the variations in an arbitrary data set.

In the work of \cite{doersch}, context prediction is used as a surrogate task. The objective for the network is to predict the relative position of two randomly sampled patches of an image. The size of the patches needs to be manually tuned such that parts of objects in the image are not over- or undersampled.

\cite{videolstm} employ a multi-layer LSTM for unsupervised image sequence prediction/reconstruction, leveraging the temporal dimension of videos as the context for individual frames.

\subsection{Generative Adversarial Nets (GANs)}%%%%%%%%%

The introduction of generative adversarial nets (GANs) \cite{gan} enabled a new discipline in unsupervised training. A generator network ($G$) and a discriminator network ($D$) are trained jointly where the $G$ tries to generate images that look as if drawn from an unlabeled data set, whereas $D$ is supposed to identify the difference between real samples and generated ones. Apart from providing compelling visual results, these networks have been shown to learn useful hierarchical representations \cite{radford16}.

\cite{igan} presents improvements in designing and training GANs, in particular, these authors achieve state-of-the-art results in semi-supervised classification on MNIST, CIFAR-10 and SVHN.

%-------------------------------------------------------------------------
%%%%%%%%% METHOD
\section{Learning by association}
A general assumption behind our work is that good embeddings will have a high similarity if they belong to the same class. We want to optimize the parameters of a CNN in order to produce good embeddings, making use of both labeled and unlabeled data.
A batch of labeled and unlabeled images ($A_{\text{img}}$ and $B_{\text{img}}$, respectively) is fed through the CNN, resulting in embedding vectors ($A$ and $B$).
We then imagine a walker going from $A$ to $B$ according to the mutual similarities, and back. If the walker ended up at the same class as he started from, the walk is correct.
The general scheme is depicted in \autoref{fig:teaser}.

\subsection{Mathematical formulation}\label{sec:formulation}
The goal is to maximize the probability for correct walks from $A$ to $B$ and back to $A$, ending up at the same class.
$A$ and $B$ are matrices whose rows index the samples in the batches.
Let's define the
\textbf{similarity between embeddings $A$ and $B$} as 
%%%%%
\begin{align}
M_{ij} \defeq A_i \cdot B_j
\end{align}
Note that the dot product could in general be replaced by any other similarity metric such as Euclidean distance. In our experiments, the dot product worked best in terms of convergence.
Now, we transform these similarities into \newline \textbf{transition probabilities from $A$ to $B$} by softmaxing $M$ over columns:
%%%%%
\begin{align}
P^{ab}_{ij} = P(B_j|A_i) \defeq& (\text{softmax}_{\text{cols}}(M))_{ij}\\
=& \exp(M_{ij}) / \sum_{j'} \exp(M_{ij'}) \nonumber
\end{align}
Conversely, we get the transition probabilities in the other direction, $P^{ba}$, by replacing $M$ with $M^T$.
We can now define the \textbf{round trip probability} of starting at $A_i$ and ending up at $A_j$:
%%%%%
\begin{align}
P^{aba}_{ij} \defeq& (P^{ab}P^{ba})_{ij} \\
=& \sum_kP^{ab}_{ik}P^{ba}_{kj} \nonumber
\end{align}

\newpage

Finally, the \textbf{probability for correct walks} becomes
%%%%%
\begin{align}
P(\text{correct walk}) = \frac{1}{|A|} \sum_{i \sim j} P^{aba}_{ij}
\end{align}
where $i \sim j \Leftrightarrow $ class($A_i$) = class($A_j$).

We define multiple losses that encourage intuitive goals. These losses can be combined, as discussed in \autoref{sec:experiments}.

\begin{align}
\mathcal{L}_{\text{total}} = \mathcal{L}_{\text{walker}} + \mathcal{L}_{\text{visit}} + \mathcal{L}_{\text{classification}}
\end{align}

\textbf{Walker loss.} The goal of our association cycles is consistency. A walk is consistent when it ends at a sample with the same class as the starting sample. This loss penalizes incorrect walks and encourages a uniform probability distribution of walks to the correct class. The uniform distribution models the idea that it is permitted to end the walk at a different sample than the starting one, as long as both belong to the same class.
The walker loss is defined as the cross-entropy $H$ between the uniform target distribution of correct round-trips $T$ and the round-trip probabilities $P^{aba}$.
%%%%%
\begin{align}
\mathcal{L}_{\text{walker}} \defeq& H(T, P^{aba})
\end{align}
with the uniform target distribution
\begin{align}
T_{ij} \defeq&  \begin{cases}
                        1/\text{$\#$class($A_i$)} \quad &\text{class($A_i$) = class($A_j$)} \\
                        0 \quad &\text{else}
                    \end{cases}
\end{align}
where $\#$class($A_i$) is the number of occurrences of class($A_i$) in $A$.

\textbf{Visit loss.} There might be samples in the unlabeled batch that are difficult, such as a badly drawn digit in MNIST. In order to make best use of all unlabeled samples, it should be beneficial to "visit" all of them, rather than just making associations among "easy" samples. This encourages embeddings that generalize better.
The visit loss is defined as the cross-entropy $H$ between the uniform target distribution $V$ and the visit probabilities $P^{\text{visit}}$. 
If the unsupervised batch contains many classes that are not present in the supervised one, this regularization can be detrimental and needs to be weighted accordingly. 
%%%%%
\begin{align}
\mathcal{L}_{\text{visit}} \defeq& H\left(V, P^{\text{visit}}\right)
\end{align}
where the visit probability for examples in $B$ and the uniform target distribution are defined as follows:
\begin{align}
P^{\text{visit}}_j \defeq& \langle P^{ab}_{ij} \rangle_i\\
V_j \defeq& 1/|B|
\end{align}

\textbf{Classification loss.} So far, only the creation of embeddings has been addressed.
These embeddings can easily be mapped to classes by adding an additional fully connected layer with softmax and a cross-entropy loss on top of the network. We call this loss classification loss. This mapping to classes is necessary to evaluate a network's performance on a test set. However, convergence can also be reached without it.

\subsection{Implementation}
The total loss $\mathcal{L}_{\text{total}}$ is minimized using Adam \cite{adam} with the suggested default settings.
We applied random data augmentation where mentioned in \autoref{sec:experiments}.
The training procedure is implemented end-to-end in TensorFlow \cite{tensorflow} and the code is publicly available.

%-------------------------------------------------------------------------
%%%%%%%%% EXPERIMENTS
\section{Experiments}\label{sec:experiments}
In order to demonstrate the capabilities of our proposed training paradigm, we performed different experiments on various data sets. Unless stated otherwise, we used the following network architecture with batch size 100 for both labeled batch $A$ (10 samples per class) and unlabeled batch $B$:
\begin{align*}
&C(32,3)\rightarrow C(32, 3)\rightarrow P(2)\\
\rightarrow \; &C(64, 3)\rightarrow C(64, 3)\rightarrow P(2)\\
\rightarrow\;  &C(128, 3)\rightarrow C(128, 3)\rightarrow P(2)\rightarrow FC(128)
\end{align*}

Here, $C(n, k)$ stands for a convolutional layer with n kernels of size $k \times k$ and stride 1.
$P(k)$ denotes a max-pooling layer with window size $k \times k$ and stride 1.
$FC(n)$ is a fully connected layer with $n$ output units.

Convolutional and fully connected layers have exponential linear units (elu) activation functions \cite{elu} and an additional L2 weight regularizer with weight $10^{-4}$ applied.

There is an additional FC layer, mapping the embedding to the logits for classification after the last FC layer that produces the embedding, i.e., $FC(10)$ for 10 classes.

\subsection{MNIST}\label{sec:mnist}
The MNIST data set \cite{mnist} is a benchmark containing handwritten digits for supervised classification.
Mutual exclusivity regularization with transformations (\cite{sajjadi2016regularization}) have previously set the state of the art among semi-supervised deep learning methods on this benchmark.
We trained the simple architecture mentioned above with our approach with all three losses from \autoref{sec:formulation} and achieved competitive results as shown in \autoref{tbl:mnist}. We have not even started to explore sophisticated additional regularization schemes that might further improve our results. The main point of these first experiments was to test how quickly one can achieve competitive results with a vanilla architecture, purely by adding our proposed training scheme. In the following, we explore some interesting, easily reproducible properties.

\begin{table*}[htb]
\begin{center}
\begin{tabular}{|c||c|c|c|}
\hline
                            & \multicolumn{3}{c|}{\# labeled samples} \\ %\cline{2-4}
Method                      & 100                  & 1000                  & All \\
\hline\hline
%Ladder, full$^\dagger$ \cite{ladder}          
                            %& 1.06 (0.37)     &   0.84 (0.08)   &   0.57 (0.02)\\
Ladder, conv small $\Gamma$ \cite{ladder}          
                            & 0.89 (0.50)     &   -   &   -\\
Improved GAN $^\dagger$ \cite{igan}                    & 0.93 (0.07)     & -                     & - \\
Mutual Exclusivity + Transform. \cite{sajjadi2016regularization}
                            &   \textbf{0.55 (0.16)}   &  -              &   \textbf{0.27 (0.02)}\\
\hline
Ours                        & 0.89 (0.08) & 0.74 (0.03) & 0.36 (0.03)\\
\hline
\end{tabular}
\end{center}
\caption{Results on MNIST. Error (\%) on the test set (lower is better). Standard deviations in parentheses.
$\dagger$: Results on permutation-invariant MNIST.}
\label{tbl:mnist}
\end{table*}

\subsubsection{Evolution of associations}
\begin{figure*}
\begin{center}
   \includegraphics[width=0.7\linewidth]{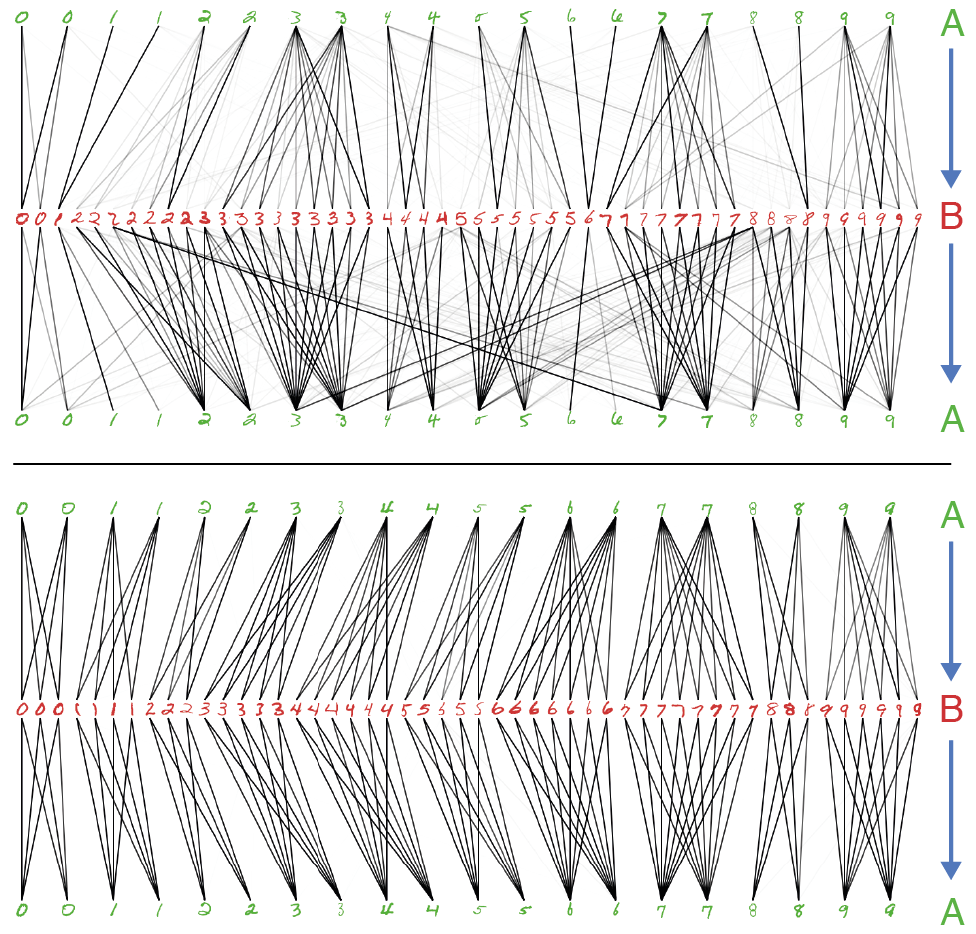}
\end{center}
   \caption{Evolution of associations. 
   Top: in the beginning of training, after a few iterations. Bottom: after convergence. Green digits are the supervised set ($A$) and red digits are samples from the unsupervised set ($B$).}
\label{fig:mnist_connections}
\end{figure*}

The untrained network is already able to make some first associations based on the produced embeddings. However, many wrong associations are made and only a few samples in the unsupervised batch ($B$) are visited: those most similar to the examples in the supervised batch ($A$).
As training progresses, these associations get better. The visit loss ensures that all samples in $B$ are visited with equal probability.
\autoref{fig:mnist_connections} shows this evolution. The original samples for a setup with 2 labeled samples per class are shown where $A$ is green and $B$ is red. Associations are made top-down. Note that the second set of green digits is equal to the first ("round-trip").
The top graphic in \autoref{fig:mnist_connections} shows visit probabilities at the beginning of training. Darker lines denote a higher probability (softmaxed dotproduct).
The bottom graphic in \autoref{fig:mnist_connections} shows associations after training has converged. This took 10k iterations during which only the same 20 labeled samples were used for $A$ and samples for $B$ were drawn randomly from the rest of the data set, ignoring labels.

\subsubsection{Confusion analysis}
\begin{figure}[h]

\begin{center}
   \includegraphics[width=1.\linewidth]{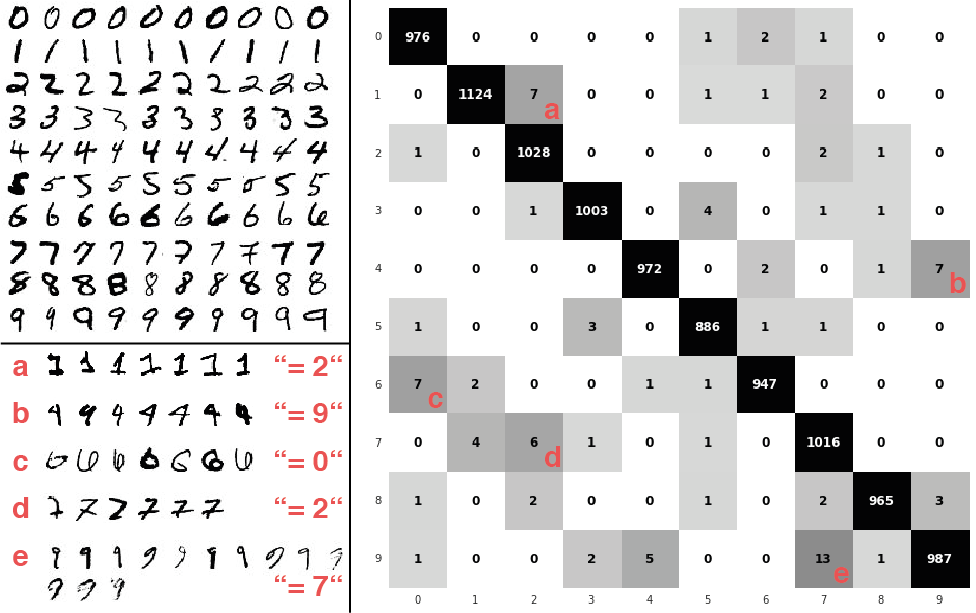}
\end{center}
   \caption{MNIST classification. Top left: All labeled samples that were used for training. Right: Confusion matrix with mistakes that were made. Test error: 0.96\%.
   Bottom left: Misclassified examples from the test.}
\label{fig:mnist_confusion}
\end{figure}

Even after training has converged, the network still makes mistakes. These mistakes can, however, be explained. \autoref{fig:mnist_confusion} shows a confusion matrix for the classification task. On the left side, all samples from the labeled set ($A$) are shown (10 per class). Those samples that are classified incorrectly express features that are not present in the supervised training set, e.g. "7" with a bar in the middle (mistaken for "2") or "4" with a closed loop (mistaken for "9").
Obviously, $A$ needs to be somewhat representative for the data set, as is usually the case for machine learning tasks.

%\subsubsection{Association as regularization}
%When we compared the performance of networks trained supervisedly versus semi-supervisedly, we found that adding our proposed method to an %existing purely supervised training scheme can act as an efficient regularizer. \autoref{fig:regularization} shows the test error curves for 20k %iterations and the histograms for the test error at the last iteration. Red and green curves are for supervised and semi-supervised training, %respectively. 
%As training data we chose 20 random subsets from MNIST for training, consisting of 100 labeled samples (10 per class).
%In the semi-supervised case, an additional 100 randomly chosen unlabeled samples are used. The results are shown in \autoref{tbl:regularization} %shows the quantitative results. Adding dropout prevented overfitting in the purely supervised case. However, adding our semi-supervised training %scheme leads to better results without the need for dropout.
%Of course, it has to be noted that the latter requires more data, which, however, doesn't need to be labeled.
%
%\begin{table}
%begin{center}
%begin{tabular}{|r|c|c|}
%\hline
% dropout rate       &   supervised      & semi-supervised \\
%\hline\hline
%0                   & 23.30 $\pm$ 15.90 & 2.21 $\pm$ 0.53 \\
%0.5                 & 17.40 $\pm$ 5.60 & 3.61 $\pm$ 0.58 \\
%0.8                 & 17.67 $\pm$ 2.76 & 2.38 $\pm$ 0.36 \\
%\hline
%\end{tabular}
%\end{center}
%\caption{Results after 20k iterations on 20 randomly chosen training sets of size 100.}
%\label{tbl:regularization}
%\end{table}
%

\subsection{STL-10} %%%%%%%%%%%%%%%%%%%%%%%%%%%%%%%%%%%%%%%%%%%%%%%%%
STL-10 is a data set of RGB images from 10 classes \cite{stl10}. There are 5k labeled training samples and 100k unlabeled training images from the same 10 classes and additional classes not present in the labeled set.
For this task we modified the network architecture slightly as follows:
\begin{align*}
&C(32,3)\rightarrow C(64, 3, \text{stride=2})\rightarrow P(3)\\
\rightarrow \; &C(64, 3)\rightarrow C(128, 3)\rightarrow P(2)\\
\rightarrow\;  &C(128, 3)\rightarrow C(256, 3)\rightarrow P(2)\rightarrow FC(128)
\end{align*}

As a preprocessing step, we apply various forms of data augmentation to all samples fed though the net. In particular, random cropping, changes in brightness, saturation, hue and small rotations.

%\begin{table}[h]
%\begin{center}
%begin{tabular}{|l|l|l|l|}
%\hline
%Method                      & Test accuracy\%  \\
%\hline\hline
%\cite{huang16} ($K$ = 64 bits)  & 76.8 ($\pm$ 0.3)\\
%Ours                            & \\
%\hline
%\end{tabular}
%\end{center}
%\caption{Results on STL-10. Standard deviation across the 10 pre-defined folds in parentheses. : add other related work results.}
%\label{tbl:stl10}
%\end{table}

We ran training using 100 randomly chosen samples per class from the labeled training set for $A$ (i.e. we used only 20\% of the labeled training images) and achieved an accuracy on the test set of 81\%.
As this is not exactly following the testing protocol suggested by the data set creators, we do not want to claim state of the art for this experiment but do consider it a promising result. \cite{huang16} achieved 76.3\% following the proposed protocol.

\begin{figure}[h]
\begin{center}
   \includegraphics[width=1.\linewidth]{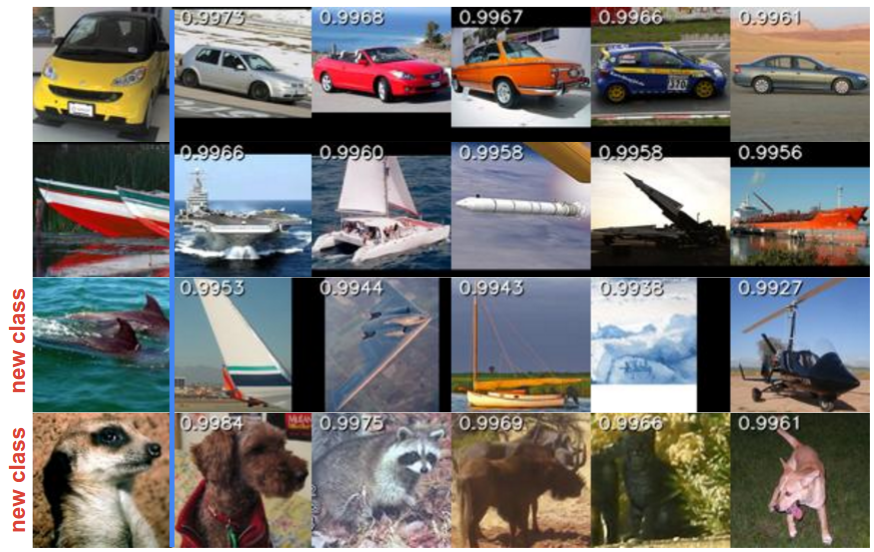}
\end{center}
   \caption{Nearest neighbors for samples from the unlabeled training set. The far left column shows the samples, the 5 other columns are the nearest neighbors in terms of cosine distance (which is shown in the top left corners of the pictures).}
\label{fig:stl10nn}
\end{figure}

The unlabeled training set contains many other classes and it is interesting to examine the trained net's associations with them. \autoref{fig:stl10nn} shows the 5 nearest neighbors (cosine distance) for samples from the unlabeled training set. The cosine similarity is shown in the top left corner of each association. Note that these numbers are not softmaxed. Known classes (top two rows) are mostly associated correctly, whereas new classes (bottom two rows) are associated with other classes, yet exposing interesting connections: The fin of a dolphin reminds the net of triangularly shaped objects such as the winglet of an airplane wing. A meerkat looking to the right is associated with a dog looking in the same direction or with a racoon with dark spots around the eyes. Unfortunately, embeddings of classes not present in the labeled training set do not seem to group together well; rather, they tend to be close to known class representations.

\subsection{SVHN} %%%%%%%%%%%%%%%%%%%%%%%%%%%%%%%%%%%%%%%%%%%%%%%%%%%%%%%
\begin{table*}[ht]
\begin{center}
\begin{tabular}{|c||c|c|c|}
\hline
                    & \multicolumn{3}{c|}{\# labeled samples} \\ %\cline{2-4}
Method               &   500                 &   1000                &   2000\\
\hline
\hline
DGN \cite{kingma}   
                    &                       &   36.02 (0.10)    & \\
Virtual Adversarial \cite{miyato2015distributional}
                    &                       &   24.63           & \\
Auxiliary Deep Generative Model \cite{maaloe2016auxiliary}
                    &                       &   22.86           & \\
Skip Deep Generative Model  \cite{maaloe2016auxiliary}    
                    &                       &   16.61 (0.24)    & \\
Imporoved GAN \cite{igan}    
                    & 18.44 (4.8)           &   8.11 (1.3)      & 6.16 (0.58) \\
Imporoved GAN (Ensemble) \cite{igan}    
                    &                       &   5.88 (1.0)      &  \\                    
Mutual Exclusivity + Transform.* \cite{sajjadi2016regularization}
                    & 9.62 (1.37)           & \textbf{4.52 (0.40)}       & \textbf{3.66  (0.14)}     \\
\hline
%Ours (sup batch size 1000)    & 8.33 (0.57)           & 7.08 (0.25)       & 5.98 (0.25) \\
%Ours (old run)                & 8.53 (0.49)           & 7.31 (0.19)       & 5.95 (0.19) \\
Ours                & \textbf{6.25 (0.32)}           & 5.14 (0.17)       & 4.60 (0.21) \\

\hline
\end{tabular}
\end{center}
\caption{
Results of comparable methods on SVHN. Error (\%) on the test set (lower is better). Standard deviations in parentheses.
\newline
*) Results provided by authors.}
\label{tbl:svhn}
\end{table*}

\begin{table*}[ht]
\begin{center}
\begin{tabular}{|c||c|c|c|c|}
\hline
\# labeled          & \multicolumn{4}{c|}{\# unlabeled samples}\\
          %\cline{2-5}
samples          &     0 &   1000 &   20000 &   all \\
\hline
\hline
 20   & 81.00 (3.01) & 81.98 (2.58) & 82.15 (1.35) & 82.10 (1.91) \\
 100  & 55.64 (6.54) & 39.85 (7.19) & 24.31 (7.19) & 23.18 (7.41) \\
 500  & 17.75 (0.65) & 12.78 (0.99) & 6.61 (0.32)  & 6.25 (0.32)  \\
 1000 & 10.92 (0.24) & 9.10 (0.37)  & 5.48 (0.34)  & 5.14 (0.17)  \\
 2000 & 8.25 (0.32)  & 7.27 (0.43)  & 4.83 (0.15)  & 4.60 (0.21)  \\
 all  & 3.09 (0.06)  & 2.79 (0.02)  & 2.80 (0.03)  & 2.69 (0.05)  \\  
\hline
\end{tabular}
\end{center}
\caption{
Results on SVHN with different amounts of (total) labeled/unlabeled training data. Error (\%) on the test set (lower is better). Standard deviations in parentheses.}
\label{tbl:svhn_semisup}
\end{table*}

The Street View House Numbers (SVHN) data set \cite{svhn} contains digits extracted from house numbers in Google Street View images. We use the format 2 variant where digits are cropped to 32x32 pixels. This variant is similar to MNIST in structure, yet the statistics are a lot more complex and richer in variation. The train and test subsets contain 73,257 and 26,032 digits, respectively.

We performed the same experiments as for MNIST with the following architecture:
\begin{align*}
&C(32,3)\rightarrow C(32, 3)\rightarrow C(32, 3)\rightarrow P(2)\\
\rightarrow \; &C(64, 3)\rightarrow C(64, 3) \rightarrow C(64, 3) \rightarrow P(2)\\
\rightarrow\;  &C(128, 3)\rightarrow C(128, 3) \rightarrow C(128, 3) \rightarrow P(2)\rightarrow FC(128)
\end{align*}

Data augmentation is achieved by applying random affine transformations and Gaussian blurring to model the variations evident in SVHN.

\subsection{Effect of adding unlabeled data}
In order to quantify how useful it is to add unlabeled data to the training process with our approach, we trained the same network architecture with different amounts of labeled and unlabeled data. For the case of no unlabeled data, only $\mathcal{L}_{\text{classification}}$ is active. In the other cases where labeled data is present, we optimize $\mathcal{L}_{\text{total}}$. We ran the nets on 10 randomly chosen subsets of the data and report median and standard deviation.

\autoref{tbl:svhn_semisup} shows results on SVHN. We used the (labeled) SVHN training set as data corpus from which we drew randomly chosen subsets as labeled and unlabeled sets. There might be overlaps between both of these sets, which would mean that the reported error rates can be seen as upper bounds.

Let's consider the case of fully supervised training. This corresponds to the far left column in \autoref{tbl:svhn_semisup}. Not surprisingly, the more labeled samples are used, the lower the error on the test set gets.

We now add unlabeled data. For a setup with only 20 labeled samples (2 per class), the baseline is an error rate of 81.00\% for 0 additional unlabeled samples. Performance deteriorates as more unlabeled samples are added. This setting seems to be pathological: depending on the data set, there is a minimum number of samples required for successful generalization.

In all other scenarios with a greater number of labeled samples, the general pattern we observed is that performance improves with greater amounts of unlabeled data. This indicates that it is indeed possible to boost a network's performance just by adding unlabeled data using the proposed associative learning scheme. For example, in the case of 500 labeled samples, it was possible to decrease the test error by 64.8\% (from 17.75\% to 6.25\%).

A particular case occurs when all data is used in the labeled batch (last row in \autoref{tbl:svhn_semisup}): Here, all samples in the unlabeled set are also in the labeled set. This means that the unlabeled set does not contain new information. Nevertheless, employing associative learning with unlabeled data improves the network's performance. $\mathcal{L}_{\text{walker}}$ and $\mathcal{L}_{\text{visit}}$ act as a beneficial regularizer that enforces similarity of embeddings belonging to the same class. This means that associative learning can also help in situations where a purely supervised training scheme has been used, without the need for additional unlabeled data.

\subsection{Effect of visit loss}
\begin{table*}[ht]
\begin{center}
\begin{tabular}{|c||c|c|c|c|}
\hline
            &  \multicolumn{4}{c|}{Visit loss weight}\\
Data set     & 0              & 0.25           & 0.5            & 1              \\
\hline
\hline
MNIST       & 5.68 (0.53)  & 1.17 (0.15) & \textbf{0.82} (0.12) & 0.85 (0.04) \\
SVHN        & 7.91 (0.40) & \textbf{6.31} (0.20) & 6.32 (0.07) & 6.43 (0.26) \\
\hline
\end{tabular}
\end{center}
\caption{
Effect of visit loss. Error (\%) on the resp. test sets (lower is better) for different values of visit loss weight. Reported are the medians of the minimum error rates throughout training with standard deviation in parentheses. Experiments were run with 1,000 randomly chosen labeled samples as supervised data set.}
\label{tbl:visitloss}
\end{table*}

\begin{table}[t]
\begin{center}
\begin{tabular}{|c|c||c|}%|c|c|}
\hline
\multirow{ 2}{*}{Data} & \multirow{ 2}{*}{Method}        & \multicolumn{1}{|c|}{Domain (source $\rightarrow$ target) }   \\
                                    && SVHN $\rightarrow$ MNIST \\%& MNIST $\rightarrow$ SVHN         & Synth $\rightarrow$ SVHN\\
\hline
\hline
\multirow{ 3}{*}{\specialcell{Source\\only}}  
& DA \cite{ganin2016}               & 45.10             \\%&               & 13.26\\
& DS \cite{bousmalis2016}           & 40.8              \\%&               & 13.3\\
& Ours                              & 18.56             \\%& 77.33         & 15.97 \\
\hline
\multirow{ 3}{*}{Adapted}       
& DA \cite{ganin2016}               & 26.15 (42.6\%)    \\%&               & 8.91 (79.7\%)\\
& DS \cite{bousmalis2016}           & 17.3 (58.3\%)     \\%&               & 8.8 (78.9\%)\\
& Ours                              & \textbf{0.51 (99.3\%)}     \\%& 33.20 (59.0\%)& \textbf{6.90 (67.6\%)} \\
\hline
\multirow{ 3}{*}{\specialcell{Target\\only}}
& DA \cite{ganin2016}               & 0.58              \\%&               & 7.80 \\
& DS \cite{bousmalis2016}           & 0.5               \\%&               & 7.6\\
& Ours                              & 0.38              \\%& 2.56          & 2.56 \\
\hline
\end{tabular}
\end{center}
\caption{
Domain adaptation. Errors (\%) on the target test sets (lower is better).
"Source only" and "target only" refers to training only on the respective data set without domain adaptation.
"DA" and "DS" stand for Domain-Adversarial Training and Domain Separation Networks, resp.
The numbers in parentheses indicate how much of the gap between lower and upper bounds was covered.}
\label{tbl:da}
\end{table}

\autoref{sec:formulation} introduces different losses. We wanted to investigate the effects of our proposed visit loss. To this end, we trained  networks on different data sets and varied the loss weights for $\mathcal{L}_{\text{visit}}$ keeping the loss weight for $\mathcal{L}_{\text{classification}}$ and $\mathcal{L}_{\text{walker}}$ constant. \autoref{tbl:visitloss} shows the results. Worst performance was obtained with no visit loss. For MNIST, visit loss is crucial for successful training. For SVHN, a moderate loss weight of about 0.25 leads to best performance. If the visit loss weight is too high, the effect seems to be over regularization of the network.. This suggests that the visit loss weight needs to be adapted according to the variance within a data set. If the distributions of samples in the (finitely sized) labeled and unlabeled batches are less similar, the visit loss weight should be lower.

\subsection{Domain adaptation}
A test for the efficiency of representations is to apply a model to the task of domain adaptation (DA) \cite{saenko2010}. The general idea is to train a model on data from a source domain and then adapt it to similar but different data from a target domain.

In the context of neural networks, DA has mostly been achieved by either \emph{fine-tuning} a network on the target domain after training it on the source domain (\cite{venugopalan, kim2014}), or by designing a network with multiple outputs for the respective domains (\cite{collobert2011, yang2016multi}), sometimes referred to as \emph{dual outputs}.

As a first attempt at DA with associative learning, we tried the following procedure that is a mix of both fine-tuning and dual outputs: We first train a network on the source domain as described in \autoref{sec:experiments}. Then, we only exchange the unsupervised data set to the target domain data and continue training. Note that here, no labels from the target class are used at all at train time.

As a baseline example, we chose a network trained on SVHN. We fed labeled samples from SVHN (source domain) and unlabeled samples from MNIST (target domain) in the network with the architecture originally used for training on the source domain and fine-tuned it with our association based approach. No data augmentation was applied.

Initially, the network achieved an error of 18.56\% on the MNIST test set which we found surprisingly low, considering that the network had not previously seen an MNIST digit. Some SVHN examples have enough similarity to MNIST that the network recognized a considerable amount of handwritten digits.

We then trained the network with both data sources as described above with 0.5 as weight for the visit loss. After 9k iterations the network reached an accuracy of 0.51\% on the MNIST test set, which is a higher accuracy than what we reached when training a network with 100 or 1000 labeled samples from MNIST (cf. \autoref{sec:mnist}). 

%A more difficult scenario is the reverse case: MNIST $\rightarrow$ SVHN. We carried out the same experiment as described above but with swapped data sets and network architectures. In order to account for the fact that some SVHN numbers are black on white background, we randomly ($p=0.5$) inverted to the MNIST samples as augmentation.

%We repeated this procedure with a more realistic setting: We trained a network on a synthetic data set of digits rendered with different fonts provided by \cite{ganin2016}. This kind of labeled data is easy and cheap to produce abundantly. We then ran domain adaptation on the much more difficult SVHN data set (without any data augmentation) and reached a test error of 6.90\% after 5k iterations.

For comparison, \cite{bousmalis2016} has been holding state of the art for domain adaptation employing domain separation networks. \autoref{tbl:da} contrasts their results with ours. Our first tentative training method for DA outperforms traditional methods by a large margin. We therefore conclude that learning by association is a promising training scheme that encourages efficient embeddings. A thorough analysis of the effects of associative learning on domain adaptation could reveal methods to successfully apply our approach to this problem setting at scale.

% formula: (source-adapted)/(source-target)

%-------------------------------------------------------------------------
\section{Conclusion}
We have proposed a novel semi-supervised training scheme that is fully differentiable and easy to add to existing end-to-end settings. The key idea is to encourage cycle-consistent association chains from embeddings of labeled data to those of unlabeled ones and back. The code is publicly available.
Although we have not employed sophisticated network architectures such as ResNet \cite{resnet} or Inception \cite{inception}, we achieve competitive results with simple networks trained with the proposed approach.
We have demonstrated how adding unlabeled data improves results dramatically, in particular when the number of labeled samples is small, surpassing state of the art for SVHN with 500 labeled samples.
In future work, we plan to systematically study the applicability of Associative Learning to the problem of domain adaptation.  
%A very promising experiment was to use our approach for domain adaptation where we outperform state of the art on SVHN $\rightarrow$ MNIST and Synthetic Digits $\rightarrow$ SVHN.
Investigating the scalability to thousands of classes or maybe even completely different problems such as segmentation will be the subject of future research.

\newpage
{\small
\bibliographystyle{ieee}
\bibliography{egbib}
}

\end{document}